\documentclass[a4paper, 10pt, conference]{ieeeconf}      

\IEEEoverridecommandlockouts                              

\usepackage{graphicx} 
\usepackage{amsmath} 

\usepackage{xcolor} 
\usepackage{verbatim}

\title{\LARGE \bf
Performance Boundary Identification for the Evaluation of Automated Vehicles using Gaussian Process Classification}

\author{Felix Batsch$^{1, 2}$, Alireza Daneshkhah$^{1}$, Madeline Cheah$^{2}$, Stratis Kanarachos$^{1}$, Anthony Baxendale$^{2}$
\thanks{$^{1}$Institute for Future Transport and Cities, Coventry University, Coventry CV1 5FB, United Kingdom}
\thanks{$^{2}$HORIBA MIRA, Nuneaton CV10 0TU, United Kingdom
        {\tt\small Corresponding Author: batschf@coventry.ac.uk}}%
\thanks{This research is funded by HORIBA MIRA}
}
\begin{document}

\maketitle
\thispagestyle{empty}
\pagestyle{empty}

\begin{abstract}

Safety is an essential aspect in the facilitation of automated vehicle deployment.  Current testing practices are not enough, and going beyond them leads to infeasible testing requirements, such as needing to drive billions of kilometres on public roads. Automated vehicles are exposed to an indefinite number of scenarios. Handling of the most challenging scenarios should be tested, which leads to the question of how such corner cases can be determined. We propose an approach to identify the performance boundary, where these corner cases are located, using Gaussian Process Classification. We also demonstrate the classification on an exemplary traffic jam approach scenario, showing that it is feasible and would lead to more efficient testing practices.

\end{abstract}

\section{INTRODUCTION}

We are seeing an increased introduction of automation technology into passenger cars and heavy goods vehicles, with the goal to ultimately reach fully automated vehicles (AVs).  AVs offer the potential for significant reductions in accident rates, emissions and traffic congestion \cite{Montanaro2018}. To that end, much research and development has gone into systems that automatically maintain adequate speed and keep within the lane by applying steering movements to aid the driver \cite{Gietelink2005}. These systems rely on information from various sensors to perceive their environment. With this information about the environment, novel methods are able to shift the driving tasks of navigation and guidance, which are traditionally performed by the human driver, to be executed by the vehicle independently.

The barrier hindering higher levels of automation (SAE levels 3-5 \cite{SAE2018}) to be introduced into public use is the assurance of their safety. In order to be accepted by the public, it needs to be validated that AVs are safe for them to use and can prevail in every situation that they encounter. Proving this with conventional testing methods would necessitate driving the AVs for hundreds of millions to billions of kilometres before rolling them out \cite{Kalra2016}.

Current validation methods range from trials on public roads over Hardware-in-the-Loop (HiL) to simulation. Especially the validation of critical driving scenarios in simulation offers cost and feasibility benefits. A common approach to validate AVs in simulation is to conduct Monte Carlo experiments, where scenarios are randomly simulated. Both, public road trials and Monte Carlo simulation suffer from the same disadvantage however, which is that most of the time spend driving on the road is uneventful \cite{Zhang2018}. Thus, a number of methods that limit the randomness and guide the search towards safety-critical scenarios have been proposed, mainly based on reducing the variation of the estimator, for example importance sampling. This accelerates the search significantly compared to Monte Carlo simulation \cite{Zhao2018}, but requires knowledge of the distribution of safety-critical scenarios which can only be taken from real world tests.

In this paper, we present a novel approach to finding safety-critical scenarios by identifying the performance boundary of an AV. The performance boundary separates the scenario space into regions according to the outcome of the scenario \cite{Mullins2018}. The outcome of a scenario can be quantitatively judged by different criticality metrics, such as the frequently used Time-to-Collision (TTC) \cite{Schreier2016}.

Along the performance boundary, small changes in the parameters that make up the scenario could result in a transition from a safe scenario to an unsafe one. These scenarios are often called \textit{corner cases}, where the individual scenario parameters are within the capability of the system, but the combination challenges the system \cite{DINSAESPEC91381-2019}. Identifying and testing for these corner case scenarios is crucial to enabling acceptable safety testing practices for AVs, and more widely, the widespread introduction of AVs \cite{Kalra2016}.

To identify the performance boundary, Gaussian Process classification (GPC) is utilised (see Section \ref{sec:3}). The GPC can probabilistically predict the outcome of scenarios that were not tested, based on known scenarios and estimate where the performance boundary is located. Due to the probabilistic nature of the Gaussian Process (GP) emulator, the proposed method also efficiently enables us to provide a point-wise confidence measure around the predicted outcomes.

The remainder of this paper is organised as follows. In Section \ref{sec:2} we review relevant literature in the field of scenario-based AV validation and supervised machine learning algorithms for classification. In Section \ref{sec:3} we formally introduce Gaussian Processes and how they are used to solve classification tasks. In Section \ref{sec:4}, the exemplary scenario considered in this work is detailed and the acquisition of the training and test data using two different sampling methods is explained. Section \ref{sec:5} shows the results of the trained GPC prediction model and the estimated performance boundary. Furthermore, we compare and analyse the differences that result from the individual data sets. The paper concludes with an outlook on future research directions in Section \ref{sec:6}.

\section{RELATED WORK}
\label{sec:2}

\subsection{Validation of AVs based on Scenarios}
Most research on the safety evaluation of AVs focuses on testing through carrying out a driving scenario involving the AV along with other actors and specified environment conditions \cite{Zlocki2015}. This stems from the problem, that the safe operation of an AV depends on the correct decision making of the AV, based on the complex interaction with other actors. This scenario-based validation is usually carried out in simulation, which offers the possibility to conduct many scenarios at low cost \cite{Khastgir2017}.

One approach to validate an AV is to conduct Monte Carlo simulations \cite{Khastgir2017}. Here, the parameters of a scenario are randomly selected from a distribution (mostly uniform distribution). After running the simulations, it is checked if the AV completed all scenarios successfully. This approach is computationally (and in case of physical testing, financially) expensive as a large number of scenarios must be run to be statistically significant. But since it achieves an even coverage of the parameter space, it is often used as baseline for comparison to other methods \cite{Althoff2011}. 

Monte Carlo methods can be improved by matching the sampling distribution to the actual distribution of the problem parameters, instead of a uniform distribution, and thus reduce the variance of the sampling. Especially importance sampling methods received a lot of attention to improve the search for scenarios that can be problematic for an AV, often called \textit{rare events} \cite{Zhang2018}. 

It was found that the evaluation time of a car following scenario can be reduced by up to a factor of 100,000, and for a lane change scenario by up to 20,000 \cite{Zhao2018, Zhao2017a}. In \cite{Gietelink2005}, importance sampling was used to evaluate an adaptive cruise control system. 

Another sampling method, Subset Simulation, was used on a lane change scenario in \cite{Zhang2018}. Subset simulation offers advantages for high dimensional stochastic models and shows a similar performance improvement as importance sampling. 

The general disadvantage of variance reduction techniques is their requirement of prior knowledge in order to shape the probability distribution used for sampling. Often this data is taken from accident databases or naturalistic driving trails \cite{Zhao2018}. This runs the risk however to exclude scenarios that emerge due to the new automation technology, which is often not considered in accident databases or naturalistic driving trails.

\subsection{Supervised Machine Learning Algorithms for Classification}

Supervised machine learning has become a topic of much discussion in recent years and they have been applied to a range of problems, from image classification to control \cite{Krizhevsky2012, Al-Qizwini2017}. 

Especially Artificial Neural Networks have received a lot of attention, leading to big improvements in their classification capabilities. They unfold their potential particularly in the area of image recognition and object classification, where large amounts of annotated data have become widely available, reducing the problem of overfitting \cite{Krizhevsky2012}. This is the crux however that makes them less attractive in the area of AV validation, where data from tests is either costly, if tests involving hardware are conducted, or computationally expensive, if simulation is involved.

Another supervised learning method is the Support Vector Machine (SVM), originally proposed in \cite{Cortes1995}. The advantage of SVMs is that they stay effective in the prediction of large dimensional data sets, even if the number of samples is small. Their efficacy in scenario prediction of human driving scenarios was shown in \cite{Remmen2018} for example. Compared to GPs, SVMs have some disadvantages however. They inherently do not provide an estimate on the confidence of the prediction, which is intrinsically provided by the GP for example. Furthermore, GPs are more flexible on custom kernel functions and learning their hyperparameters from data \cite{Rasmussen2006}. 

The k-Nearest Neighbour (k-NN) algorithm can also be used to classify data \cite{Altman1992}. The k-NN algorithms predict new data points from known data in the vicinity of the predicted point. This makes it unreliable however, if data points close to or on the boundary between classes should be predicted, which is critical to finding corner case scenarios for the evaluation of automated vehicles.

\section{GAUSSIAN PROCESSES}
\label{sec:3}

Gaussian Processes are a class of supervised machine learning algorithms, that describe a functional relation as a multivariate Gaussian distribution and can thus be used for non-linear regression and classification problems \cite{Rasmussen2006}. They have been used to model and predict trajectories of vehicles and pedestrians, as shown in \cite{Kim2011} and \cite{Ellis2009a}. Furthermore, GPs have been used to model the driving intention of human drivers, particularly for scenarios at intersections \cite{Tran2013, Armand2013}. 

Gaussian Processes have been relatively unexplored in the context of performance classification of AVs. In \cite{Mullins2018} Gaussian Process Regression (GPR) is used to adaptively search the state space of an autonomous, unmanned underwater vehicle. The authors also define a performance boundary, where the performance of the system transitions from one mode to another, due to changes in the environment. The concept of a performance boundary is further developed and adapted to ground AVs in this paper. 

An application of GPR in an automotive context is presented in \cite{Huang2017}, where the GPR was used to estimate the probability distribution of a scenario, which was subsequently used for importance sampling. They study their procedure on a lane change scenario and show an improvement of the evaluation effort compared to crude Monte Carlo sampling.

The GPC in this application is trained on data obtained from a traffic jam approach scenario, which is further described in Section \ref{sec:4}. The scenarios were executed using the vehicle simulation software CarMaker \cite{IPG2019}, and for the purpose of this paper, the simulation is regarded as ground truth. For this to be valid in the overall scope of validating the AV, physical tests have to follow up and validate the simulation.

\subsection{Formal Description of Gaussian Processes}

The GP model used in this paper is based on the extensive work of \cite{Rasmussen2006}. The formal definition of a Gaussian Process is denoted by a prior distribution

\begin{equation}
    f(\boldsymbol{x}) \sim \mathcal{GP}(m(\boldsymbol{x}),k(\boldsymbol{x},\boldsymbol{x'}))   \,,
    \label{eq:1}
\end{equation}

with a mean of $m(\boldsymbol{x})$ and the kernel function $k(\boldsymbol{x},\boldsymbol{x'})$. The kernel function used in this paper is the radial basis function (RBF), also known as squared exponential kernel \cite{Rasmussen2006}.

We consider a data set $\boldsymbol{\mathcal{D}} = \{ (\boldsymbol{x}_i, y_i) | i = 1,\dots,n \}$, consisting of $n$ samples, wherein $\boldsymbol{x}_i$ denotes the vector of input data taken from the input space $\mathcal{X}$, and $y_i = f(\boldsymbol{x}_i)$ the corresponding output observations. With the definition of a Gaussian Process from Eq. (\ref{eq:1}), we can describe a joint prior distribution for the observed outputs $\boldsymbol{f}$ and the predicted outputs $\boldsymbol{f}_*$:

\begin{equation}
    \begin{bmatrix}
        \boldsymbol{f} \\[0.3em]
        \boldsymbol{f_*}
    \end{bmatrix}
    \sim \mathcal{N}
    \begin{pmatrix}
        0, &
        \begin{bmatrix}
           K    & K_*^T      \\[0.3em]
           K_*  & K_{**}
        \end{bmatrix}
    \end{pmatrix}   \,,
    \label{eq:2}
\end{equation}

with an assumed mean of zero and the covariance matrices\\
$K=k(\boldsymbol{X},\boldsymbol{X})$\\
$K_*^T=k(\boldsymbol{X},\boldsymbol{X}_*)$\\
$K_*=k(\boldsymbol{X}_*,\boldsymbol{X})$\\
$K_{**}=k(\boldsymbol{X}_*,\boldsymbol{X}_*)$\\
for all observed and predicted data points. Here $\boldsymbol{X}$ denotes a $d\times n$ matrix of the training inputs $\{\boldsymbol{x}_{i}\}_{i=1}^{n}$ (also known as the design matrix), $d$ stands for the dimension of input space $\mathcal{X}$, and $\boldsymbol{X}_*$ is the matrix of test inputs. The subscript $*$ differentiates the test/predicted data from the observed data. To simplify the problem, the mean function is usually assumed to be zero, which does not limit the mean of the posterior to zero.

To obtain the posterior distribution over the predicted value we can condition the joint prior distribution to

\begin{equation}
    \boldsymbol{f_*}|\boldsymbol{f},X,X_* \sim \mathcal{N} (K_*K^{-1}\boldsymbol{f}, K_{**}-K_*K^{-1}K_*^T) \,.
    \label{eq:3}
\end{equation}

The predicted value $\boldsymbol{f_*}$ and its uncertainty are thus given by the mean and covariance of the posterior distribution evaluated over $\boldsymbol{X_*}$.

\subsection{Gaussian Process Classification}
Considering a new data set $\boldsymbol{\mathcal{D}} = \{ (\boldsymbol{x}_i, y_i) \}$, wherein $y_i$ now describes a number of discrete class labels according to $\boldsymbol{x}_i$. GPs can be used to solve classification problems by giving predictions in from of class probabilities $y_*$. This is done by squashing the output of a regression model through a logistic function (e.g. sigmoid function, $\sigma(\cdot)$) to transform it from a domain of $(-\infty,\infty)$ to $[0,1]$.

The classification is done in two steps, by first predicting a latent variable $f_*$ corresponding to an input value $\boldsymbol{x}_*$

\begin{equation}
    p(f_*|\boldsymbol{X},\boldsymbol{y},\boldsymbol{x}_*) = \int p(f_*|\boldsymbol{X},\boldsymbol{x}_*,\boldsymbol{\mathrm{f}}) p(\boldsymbol{\mathrm{f}}|\boldsymbol{X},\boldsymbol{y}) \, \mathrm{d}\boldsymbol{\mathrm{f}} \,.
    \label{eq:7}
\end{equation}

The probabilistic prediction can then be calculated in the second step using 

\begin{equation}
    p(y_*|\boldsymbol{X},\boldsymbol{y},\boldsymbol{x}_*) = \int \sigma(f_*) p(f_*|\boldsymbol{X},\boldsymbol{y},\boldsymbol{x}_*) \, \mathrm{d}f_* \,.
    \label{eq:8}
\end{equation}

The likelihood function in Eq. (\ref{eq:7}) is non-gaussian due to the discrete class labels in $y$. Therefore, a computationally feasible Laplace approximation must be used to approximate the integral. 
\par
In this paper, we predict AV scenario outcomes using GPC. In the remaining sections we describe this application of GPC on an exemplary scenario.

\section{DATA ACQUISITION THROUGH SIMULATION}
\label{sec:4}

\begin{figure}[b]
    \centering
    \parbox{3in}{\includegraphics[width=3in]{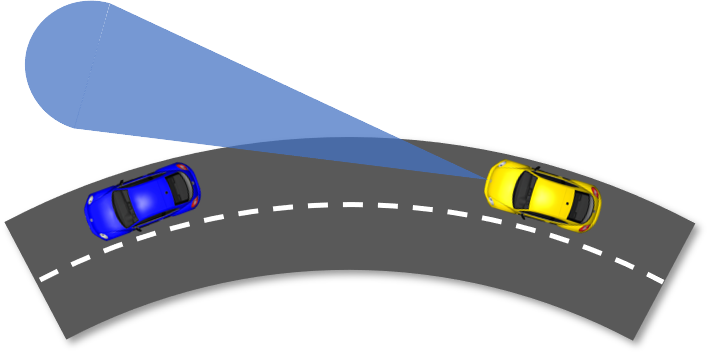}}
    \caption{Traffic jam approach scenario with the blue target vehicle and the yellow ego vehicle equipped with a radar sensor}
    \label{fig:scene}
\end{figure}

The GPC was trained on data obtained from simulations conducted with the CarMaker simulation software \cite{IPG2019}. We consider an exemplary scenario where an AV approaches a traffic jam in which vehicles are moving considerably slower than the approaching ego vehicle. Additionally, the end of the traffic jam is situated in a left turn of a curve with a radius of 50 m. To simplify the simulation, only the last vehicle in the traffic jam is modelled, and there are no additional obstructions between the ego vehicle and the traffic jam. The system under test is an automated vehicle equipped with a radar sensor to detect objects within its sensing arc. Sensor uncertainty was neglected here but will be considered in future work. If a vehicle is detected, the longitudinal control initiates a braking manoeuvre in order to avoid a collision. The scenario is illustrated in Fig. \ref{fig:scene}. The preset vehicle and radar sensor models of CarMaker were used in this work.

Three parameters were considered in this paper: the speed of the approaching ego vehicle (\textit{speed ego}), the speed of the last vehicle in the traffic jam (\textit{speed target}), and the aperture angle of the radar sensor of the ego vehicle (\textit{aperture angle}). For feasibility purposes, the variation of the scenario has been restricted to these three parameters but adding more parameters, such as additional sensors and perception uncertainties, is in the scope of our future research. The parameters are varied within fixed ranges, which can be found in Table \ref{tab:1}. We note that excluding extreme parameter combinations a priori is avoided, as this would make a possibly invalid assumption and exclude potential corner case scenarios.

\begin{table}[t]
    \caption{Simulation Parameters and Ranges}
    \label{tab:1}
    \normalsize
    \begin{center}
        \begin{tabular}{|c|c|c|}
            \hline
            \textbf{Simulation Parameters} & \textbf{Lower limit} & \textbf{Upper limit}\\
            \hline
            speed ego [km/h] & 40 & 70\\
            \hline
            speed target [km/h] & 5 & 20\\
            \hline
            aperture angle [deg] & 10 & 25\\
            \hline
        \end{tabular}
    \end{center}
\end{table}

The outcome of the simulated scenarios was evaluated on whether the AV was able to prevent a collision or not, resulting in a binary classification problem. It should be mentioned at this point however, that GPC can also be used for multi-class classification problems, in case multiple events should be considered and this is something we also consider for future work (see Section~\ref{sec:6}). 

Two sampling methods were used to create data sets of different size in order to compare the efficacy of the sampling methods, and the amount of necessary training data. The data sets were obtained through Monte Carlo sampling and the Minimax Latin Hypercube design method. The rationale in including different sampling methods is explained in the following.

\subsection{Monte Carlo Sampling}
Baseline data sets to train and test the GPC prediction model were generated by sampling from a uniform distribution, confined to the limits described in Table \ref{tab:1}. The limits were chosen on representative values for the scenario, to skew the parameter space and exclude parameter combinations of no interest.

In the applied Monte Carlo sampling, every parameter had its own independent distribution. Monte Carlo sampling does not guarantee that the samples are homogeneously spread throughout the parameter space but is included here as it is frequently used as basis for comparison \cite{Althoff2011}.

\subsection{Latin Hypercube Sampling}

Furthermore, data sets were created using Latin Hypercube (LHC) sampling \cite{McKay1979}. LHC ensures that the parameter space is evenly covered by dividing each parameter into intervals which are then randomised. This reduces the risk that large areas of the parameters space remain uncovered or that samples are too close together.

The generated data sets were split into a training and a test set. The training set was used to tune the hyperparameters of the GPC. The different data sets and their sizes can be found in Table \ref{tab:2}.

\begin{table}[t]
    \caption{Data sets}
    \label{tab:2}
    \normalsize
    \begin{center}
        \begin{tabular}{|c|p{4.8em}|p{5em}|p{5em}|}
            \hline
            \textbf{Name} & \textbf{Sampling Method} & \textbf{Size \newline training set} & \textbf{Size \newline test set}\\
            \hline
            MC100 & Monte Carlo & 90 & 10\\
            \hline
            MC1000 & Monte Carlo & 900 & 100\\
            \hline
            LHC100 & Latin Hypercube & 90 & 10\\
            \hline
            LHC1000 & Latin Hypercube & 900 & 100\\
            \hline
        \end{tabular}
    \end{center}
\end{table}

\section{RESULTS}
\label{sec:5}

\subsection{Finding the Performance Boundary}

\begin{figure}[b]
    \centering
    \parbox{3in}{\includegraphics[width=3in]{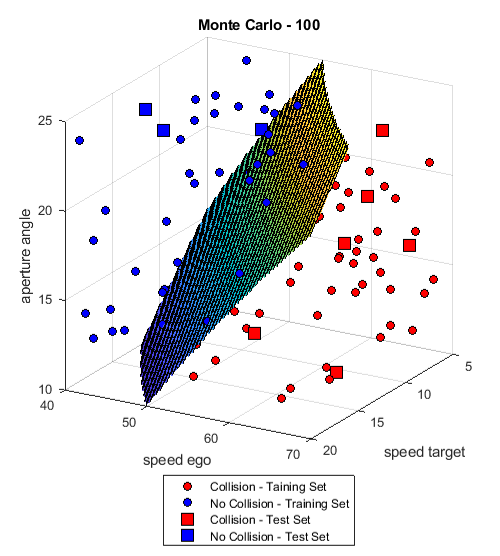}}
    \caption{Boundary estimation based on the MC100 data set}
    \label{fig:MCboundary}
\end{figure}

Separate Gaussian Process classifications were trained and tested on the four data sets described in Table \ref{tab:2}. For the scenario described in Section \ref{sec:4}, a clear performance boundary was found, which separates the scenarios that ended successfully, i.e. the AV could prevent a collision, from the ones that were unsuccessful and ended in a collision. The performance boundary along with the training and test data of the MC100 data set is visualised in Fig. \ref{fig:MCboundary}.

\begin{figure}[t]
    \centering
    \parbox{3in}{\includegraphics[width=3in]{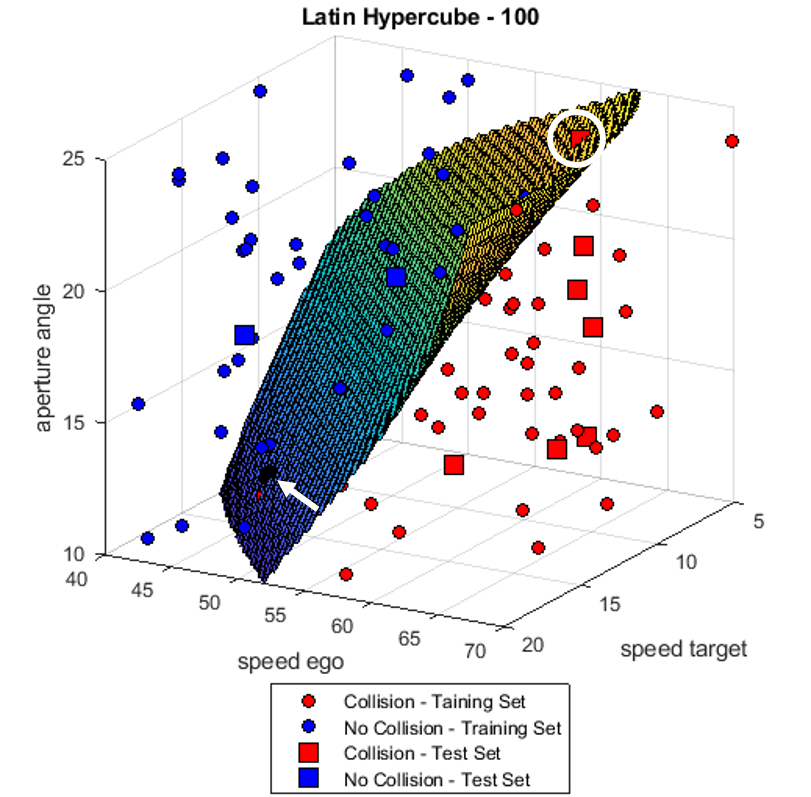}}
    \caption{Boundary estimation based on the LHC100 data set with a single misclassification marked by a white circle and an exemplary scenario on the performance boundary indicated by a black dot (white arrow)}
    \label{fig:LHCboundary}
\end{figure}

While the MC100 data set yielded good classification results, with no misclassifications on the test set, there is the possibility that a misclassification occurs, if a test data point is very close to the performance boundary and thus not included in its prediction. For the LHC100 test data set one misclassification occurred, where a test data point was on the performance boundary, indicated by the white circle in Fig. \ref{fig:LHCboundary}. If this data point would be included in the training set, the performance boundary would have shifted accordingly. 

One exemplary scenario on the performance boundary is depicted as a black dot in Fig. \ref{fig:LHCboundary} (white arrow). The explicit parameter values of this predicted scenario on the performance boundary and the two closest, simulated scenarios on either side of the boundary are listed in Table \ref{tab:3}.

\begin{figure}[b]
    \centering
    \parbox{3in}{\includegraphics[width=3in]{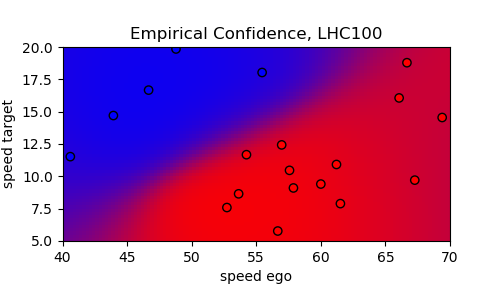}}
    \caption{Empirical confidence provided by the Gaussian Process classification at a constant aperture angle of 17.5$^\circ$. The data points displayed are in the range [16, 19]$^\circ$.}
    \label{fig:LHCuncertainty100}
\end{figure}

Furthermore, it can be seen that training the GPC on the two different data sets yields slightly different performance boundary estimates. This is due to the sparsity and different distributions of data in the sets and leads to a maximum Euclidean distance of 4.5 between the two performance boundary estimates. This deviation of the estimated performance boundary can be minimised by increasing the number of data points. The GPC on the MC1000 and LHC1000 data sets with 1000 data points reduced the maximum Euclidean distance between the performance boundary estimates to 1.25. 

Additionally, we analysed which sampling method yields a better estimation of the performance boundary on sparse data. The Latin Hypercube sampling method had a slightly better estimation, as the maximum Euclidean distance between the performance boundary estimations of the LHC100 and LHC1000 data set was 3.75. The maximum Euclidean distance between the performance boundary estimations of the MC100 and the MC1000 data set was found to be 4.75.

\begin{table}[t]
    \caption{Exemplary scenarios on the performance boundary \newline (LHC100 data set)}
    \label{tab:3}
    \normalsize
    \begin{center}
        \begin{tabular}{|p{4.2em}|p{5.2em}|p{4.8em}|p{5.3em}|}
            \hline
            \textbf{speed ego [km/h]} & \textbf{speed target [km/h]} & \textbf{aperture angle [deg]} & \textbf{Scenario Outcome}\\
            \hline
            47.27 & 15.76 & 11.36 & Collision\\
            \hline
            46.97 & 15.30 & 13.33 & No-Collision\\
            \hline \hline
            47.25 & 15.5 & 12.25 & Boundary\\
            \hline
        \end{tabular}
    \end{center}
\end{table}

\subsection{Uncertainty Measure of the Classification}

An advantage of the GPC is that it provides a measure on the uncertainty of the prediction model. Since the prediction is probabilistic, we can calculate the empirical confidence of estimations, which gives additional information on the prediction.

In Fig. \ref{fig:LHCuncertainty100} the prediction output of the GPC model trained on the LHC100 data set is shown for a constant aperture angle of 17.5$^\circ$. Also displayed are the data points that have the biggest influence on the estimation of the performance boundary. These are the data points located in the vicinity of the constant third parameter; data points with an aperture angle on the interval [16,19]$^\circ$. As can be seen from Fig. \ref{fig:LHCuncertainty100}, the performance boundary is fuzzy, especially in the areas close to the limits of the data set. This gives a direct indication on the confidence level of predictions from the GPC model.

\begin{figure}[b]
    \centering
    \parbox{3in}{\includegraphics[width=3in]{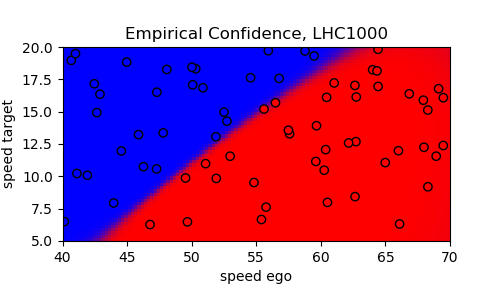}}
    \caption{Empirical confidence provided by the Gaussian Process classification at a constant aperture angle of 17.5$^\circ$. The data points displayed are in the range [17, 18]$^\circ$.}
    \label{fig:LHCuncertainty1000}
\end{figure}

With an increased number of data points, especially close to the true location of the performance boundary, the empirical confidence is increased and the performance boundary can be drawn much sharper. Fig. \ref{fig:LHCuncertainty1000} shows the empirical confidence of the LHC1000 data set. It is clearly visible that the increased number of data points, which are here on an interval of [17,18]$^\circ$, increase the confidence of the prediction along the performance boundary area. Similar to Fig. \ref{fig:LHCuncertainty100} the empirical confidence decreases towards the limits of the data set, due to missing data beyond the limit. 

\section{CONCLUSIONS AND FUTURE WORK}
\label{sec:6}

In this work, we applied Gaussian Process classification to the problem of automated vehicle validation. It was found that the performance boundary, which separates successful from unsuccessful scenarios, can be adequately estimated from simulated data.

Knowledge about the location of the performance boundary can be used to identify corner cases, which are the scenarios on the performance boundary. We detailed the predicted parameters of such a scenario in Table \ref{tab:3}. These scenarios can be used to test the automated vehicle and help to identify erroneous behaviour or limitations in the design of the systems.

It was shown that a Latin Hypercube design of the data set yields a better prediction than randomised Monte Carlo sampling when looking at sparse data sets, an advantage which decreases, however, with larger data sets. Furthermore, the amount of available data around the performance boundary naturally improves the confidence of the prediction model on the location of the performance boundary. 

The disadvantage of Gaussian Process classification lies in its lack of scalability, as it scales with $\mathcal{O}(n^3)$. High dimensionality of the input might thus necessitate dimensionality reduction methods such as sparse Gaussian Processes to stay computationally viable \cite{Lawrence2002}.

In future work, we plan to integrate the prediction model in an adaptive framework, to concentrate the sampling around the performance boundary and thus reduce the necessary simulations to find the performance boundary. Additionally, we plan to extend the model to include a multi-dimensional parameter space and multiple criticality classes, which the scenarios are classified in. These could for example include a classification of near miss scenarios. Further investigation is also necessary to look at the influence of different kernel functions and their impact on the classification.

\addtolength{\textheight}{-12cm}   








\end{document}